\documentclass{article}
\usepackage{spconf,amsmath,graphicx,booktabs,multirow,array,url,float}
\usepackage{fontspec}
\usepackage{xeCJK}
\usepackage[hidelinks]{hyperref}
\setCJKsansfont[
  Path=fonts/,
  AutoFakeBold=2,
  AutoFakeSlant=0.2
]{NotoSerifCJKsc-Regular.otf}
\setCJKmonofont[
  Path=fonts/,
  AutoFakeBold=2,
  AutoFakeSlant=0.2
]{NotoSerifCJKsc-Regular.otf}

\title{ASR-Roundtrip Evaluation Can Mask Context- and Convention-Dependent Reading Errors in Chinese News TTS}

\name{Shijun Luo\sthanks{Corresponding author.} and Lizhi Wan}
\address{NetEase Cloud Music, Hangzhou, China}

\begin{document}
\maketitle

\begin{abstract}
ASR-roundtrip evaluation is widely used as a scalable proxy for text-to-speech (TTS) intelligibility, but it can produce false negatives for reading errors perceived by listeners. We study Chinese news TTS spans whose correct reading depends on context or domain conventions, such as sports scores, aircraft models, technical units, and membership names. In these cases, Raw TTS can choose a plausible but wrong reading while ASR transcribes the audio as the intended or surface-correct text. A targeted audit over 110 high-risk MiMo TTS cases, reported with a complete denominator, confirms 46 masked false negatives, 9 exposed TTS errors, and 55 cases with no Raw TTS error. A span-isolation diagnostic re-exposes 18/46 previously masked errors. A Raw-only CosyVoice audit on the same targeted pool confirms 51 masked cases. Across the 97 TTS-specific audio files labeled confirmed masked across the two audits, Qwen3-ASR surface-recovers 40 cases, whereas Paraformer does so in only 2. The results suggest that ASR-roundtrip is useful for screening but insufficient as standalone ground truth for Chinese news reading-risk evaluation.
\end{abstract}


\section{Introduction}

Automatic speech recognition (ASR) roundtrip evaluation is attractive for TTS: synthesize speech, transcribe it, and compare the transcript to a reference. It scales cheaply and often tracks intelligibility, but its reliability depends on the task and protocol~\cite{taylor2021asr,favre2013asr}. This paper shows a specific failure mode for Chinese news TTS: the audio can be wrong to a listener, while ASR transcribes it as the intended or surface-correct text.

The problem appears in short, high-prior written forms. In a snooker report, \emph{13-11} should be read as a score, i.e., 十三比十一, rather than a range-like reading such as 十三至十一. In a military report, 伊尔-76 is an aircraft model, not a negative number. In technology and automotive news, \emph{640kW}, \emph{350Wh/kg}, or \emph{88VIP} require domain conventions. These errors are often fluent: a listener hears the wrong reading, but the transcript can still look correct.

We frame the core family as \emph{Context-Dependent Reading Decisions} (CDRD): written spans whose correct reading depends on context beyond the local character sequence. We also include CDRD-adjacent risks such as unit strings, membership names, abbreviations, and foreign names. These adjacent cases are not always CDRD under the strict definition, but they expose the same evaluation failure: ASR can normalize or recover a surface-correct transcript despite the wrong spoken reading.

We study an evaluation failure rather than propose a new TTS frontend. We provide: (i) a targeted audit, reported with a complete denominator, for TTS-wrong / ASR-surface-correct false negatives; (ii) a span-isolation diagnostic showing that removing sentence context re-exposes many errors; (iii) cross-TTS and cross-ASR controls showing both recurrence and evaluator dependence; and (iv) a human-audited evaluation protocol.

\section{Reading-Risk Benchmark}

\subsection{Risk definition}

A CDRD span is a written span $x$ whose correct reading depends on context $c$. The same local pattern can require different readings across domains: a hyphen may mark a score, range, model, or subtraction-like form; a mixed digit-letter span may be a unit, product name, or membership tier. CDRD overlaps with Chinese text normalization and G2P/polyphone disambiguation~\cite{dai2022flattn,choi2021g2p}, but our target is the downstream evaluation failure when audio with a wrong reading is recovered as surface-correct text. We use three case labels:
\begin{itemize}
\item \textbf{CDRD-entity}: scores, military/aircraft models, financial quarters, generation labels, and entity-dependent hyphenated forms.
\item \textbf{CDRD-polyphone}: Chinese polyphone ambiguity, tracked separately because it is closer to G2P.
\item \textbf{CDRD-adjacent}: units, percentages, mixed-script strings, memberships, abbreviations, and foreign names.
\end{itemize}

Each case is annotated at the risk-span level with surface form, expected reading, known negative readings when available, context evidence, and whether transcript-based scoring is allowed.

\subsection{Data construction}

We start from 108,124 company-produced Chinese news scripts used in a production TTS workflow. Taxonomy-driven mining follows the text-normalization tradition of explicitly modeling non-standard written forms~\cite{sproat2001normalization,ebden2015kestrel}: hyphens, scores, mixed digit/Latin strings, technical units, financial quarters, generation labels, aircraft/military models, and known polyphone-prone spans. For real news, a scripted candidate-construction pipeline cleans titles and summaries, filters missing or unusable titles and rows without rule-detected auditable risk spans, scores candidates by the number and diversity of risk spans, and applies type/domain caps to select a 500-case real-news candidate pool. These construction filters use only text and risk-span metadata before TTS/ASR generation, so they are risk-coverage and data-availability filters rather than outcome filters.

We also build a 5K synthetic hard-case pool and freeze a 200-case benchmark for both human listening audits and automatic TTS/ASR experiments. The frozen set contains 155 real-news cases and 45 synthetic hard cases. It covers all three labels and seven news domains: 85 CDRD-entity, 35 CDRD-polyphone, and 80 CDRD-adjacent/non-CDRD cases; domains include auto (44), general (43), finance (33), tech (31), sports (27), international (12), and military (10). Human listening audits are conducted after the benchmark and targeted audit pools are frozen; human judgments of TTS or ASR outcomes are not used to select the candidate pool.

\section{Evaluation Protocol}

\subsection{Systems}

The primary system pairs MiMo-V2.5-TTS API synthesis with the audio-capable MiMo \texttt{mimo-v2.5} API~\cite{xiaomi_mimo_api} prompted for verbatim transcription; \texttt{mimo-v2-omni} is used as a fallback and protocol-ablation route. The strict ASR protocol mainly targets spoken numeral preservation and discourages transcript normalization; it reduces but cannot fully suppress unit- or entity-level canonicalization in the decoder or post-processing. We use Whisper-small~\cite{radford2023whisper} as a non-MiMo ASR control and CosyVoice-300M-SFT~\cite{du2024cosyvoice} as a Raw-only second-TTS validation. We additionally transcribe the 220 TTS-specific Raw files from the two 110-case targeted audits and the 46 aligned MiMo clips with Paraformer-zh v2.0.4~\cite{gao2022paraformer} and Qwen3-ASR-1.7B~\cite{shi2026qwen3asr}. Paraformer uses FSMN-VAD without punctuation, hotwords, or an external language model; Qwen uses its local Transformers backend with empty context and automatic language detection. Neither system receives source text or target readings.

We evaluate two text conditions. In the Raw condition, the TTS input is the original news title and summary, without rewriting CDRD/CDRD-adjacent risk spans into expected spoken forms; model settings are matched across conditions, and no target-span expected readings are supplied in Raw. In the Structured condition, the same source text is converted into a protocol-level spoken-text form by instantiating precomputed risk-span annotations: expected readings replace written spans, and negative readings and context evidence are retained for audit. Because the readings come from predefined rules and mappings that also define the benchmark targets, Structured is an oracle-style diagnostic condition. It should be interpreted as an upper bound, not as a deployable frontend or a fair end-to-end system baseline.

\subsection{Human span audit}

One primary annotator listened to Raw and Structured audio for all 200 frozen benchmark cases and counted correct risk spans. This is a targeted span-audit task, not a blind MOS or blind preference test. The review page shows the original text, Structured TTS text, key risk spans with expected and known negative readings, Raw and Structured audio players, and ASR transcripts.

The annotation protocol is audio-first. For the primary pass and IAA passes, annotators judge risk-span correctness from the audio against expected and known negative readings before using ASR transcripts as audit context. ASR transcripts are not treated as evidence that audio is correct.

The primary pass covers all 200 cases. A 30-case IAA subset is independently reviewed by two additional annotators, stratified as 10 CDRD-entity, 10 CDRD-polyphone, and 10 CDRD-adjacent/non-CDRD cases. Over $N$ cases, the metric is case-macro risk-span audio accuracy for condition $s$:
\[
\mathrm{Acc}_{s} =
\frac{1}{N}\sum_{i=1}^{N}
\frac{n^{\mathrm{correct}}_{i,s}}
{n^{\mathrm{risk}}_{i}},
\quad s \in \{\mathrm{Raw}, \mathrm{Structured}\}.
\]
If an entered count exceeds the number of spans, the official score clamps it to the total and preserves the original value for audit. We report correct-count IAA rather than better-pipeline agreement because the latter is more subjective.

\subsection{Targeted masked-error audit}

We conduct a 110-row targeted audit over high-risk frozen-benchmark candidates. A row enters the audit pool as a candidate, not as a confirmed masked case. Candidates are mined using high-risk span rules and suspected masking signals, including CDRD/CDRD-adjacent forms, Raw/Structured disagreements, and ASR transcripts that appear surface-correct despite high-risk audio.

The masked-error audit is also audio-first. The primary reviewer first listens to Raw TTS and judges whether the target span is spoken incorrectly. Only after this audio judgment does the reviewer consult ASR transcripts to categorize the row as:
\begin{itemize}
\item \textbf{confirmed masked}: Raw TTS is wrong and at least one ASR route writes the expected/surface-correct form;
\item \textbf{exposed TTS error}: Raw TTS is wrong and ASR exposes or preserves the wrong reading;
\item \textbf{no Raw TTS error}: Raw TTS is correct;
\item \textbf{uncertain/not judgeable}: insufficient evidence.
\end{itemize}
The final 110 audit categories are then fully checked by a second reviewer before the counts in Table~\ref{tab:audit} are reported. As an additional robustness check, we build a 30-row label-blind relabel subset stratified from the targeted audit pool, including 15 confirmed-masked cases that are not re-exposed by span isolation, 7 exposed-error cases, and 8 no-error cases. Blind relabeling reaches 23/30 exact agreement over full audit labels ($\kappa=0.634$), and 27/30 agreement for confirmed-masked versus other outcomes ($\kappa=0.800$). This second pass and blind subset cover audit categories, while the 30-case IAA above covers correct-count labels for the 200-case Raw/Structured span audit.
This targeted audit demonstrates the mechanism, not its production prevalence.

\section{Results}

\subsection{ASR false negatives in targeted audits}

Table~\ref{tab:audit} reports the targeted audit, with a complete denominator, that supports the ASR false-negative claim. In the MiMo audit, human listening confirms 46 cases where Raw TTS is wrong and ASR transcribes the audio as the expected or surface-correct text. The same audit explicitly accounts for 9 exposed TTS errors and 55 candidates with no Raw TTS error, avoiding positive-only reporting. The CosyVoice column uses the same targeted pool and is interpreted in Section~4.5 as cross-TTS mechanism validation, not as a second prevalence estimate.

\begin{table}[H]
\centering
\caption{Primary targeted-audit evidence. Counts are targeted audit yields, not production prevalence.}
\label{tab:audit}
\small
\begin{tabular}{@{}lrr@{}}
\toprule
Outcome & MiMo Raw & CosyVoice Raw \\
\midrule
confirmed masked & 46 & 51 \\
exposed TTS error & 9 & 27 \\
no Raw TTS error & 55 & 30 \\
uncertain / not judgeable & 0 & 2 \\
total targeted pool & 110 & 110 \\
\bottomrule
\end{tabular}
\end{table}

This evidence can be read as an ASR false-negative contingency table: wrong Raw audio plus surface-correct ASR is a masked false negative (46); wrong Raw audio plus exposed ASR is an exposed TTS error (9); correct Raw audio is no Raw TTS error (55). Masking is protocol-sensitive: a conservative separator-normalized match against either the written target span or the expected reading identifies surface-correct recovery in 34/46 cases under default ASR and 27/46 under the MiMo-V2-Omni strict route; normalization removes only whitespace, hyphens, colons, full-width colons, and middle dots. Thus the audit label is not defined by one ASR prompt alone; it requires a human-confirmed wrong Raw reading and at least one surface-correct ASR route. By benchmark label, the MiMo outcomes are 33/6/42 for CDRD-entity, 3/1/7 for CDRD-polyphone, and 10/2/6 for CDRD-adjacent/non-CDRD, respectively ordered as confirmed masked, exposed error, and no Raw TTS error. The denominator is targeted and high-risk, so it supports existence and mechanism, not incidence.

\subsection{Representative failure modes}

Table~\ref{tab:examples} makes the masking mechanism concrete: the Raw audio contains the wrong reading, while ASR recovers a surface-correct or conventional written form. The route is not uniform: scores and model names mainly reflect contextual recovery, whereas units and membership names often reflect convention- or normalization-driven canonicalization. The 46 MiMo confirmed masked cases are not random acoustic glitches. They cover sports scores (20), kW/kWh unit strings (10), military models (9), hyphen-range or torque forms (3), and smaller financial-quarter, compute-unit, membership, and voltage categories (one each).

\begin{table}[H]
\centering
\caption{Representative confirmed masked cases.}
\label{tab:examples}
\scriptsize
\setlength{\tabcolsep}{2.5pt}
\renewcommand{\arraystretch}{0.92}
\begin{tabular}{@{}>{\raggedright\arraybackslash}p{0.14\columnwidth}
                    >{\raggedright\arraybackslash}p{0.22\columnwidth}
                    >{\raggedright\arraybackslash}p{0.23\columnwidth}
                    >{\raggedright\arraybackslash}p{0.31\columnwidth}@{}}
\toprule
Surface & Intended & Raw heard & ASR recovery / route \\
\midrule
\emph{13-11} & 十三比十一 & 十三至十一 & score form (context) \\
\emph{F-35} & F三五 & F杠三五 & \emph{F-35} (model) \\
伊尔-76 & 伊尔七六 & 伊尔负七六 & aircraft form (entity) \\
\emph{88VIP} & 八十八VIP & 八十八伏IP & \emph{88VIP} (membership) \\
\emph{350Wh/kg} & 三百五十瓦时每公斤 & partial unit-letter reading & normalized unit \\
\bottomrule
\end{tabular}
\end{table}

\subsection{Context isolation}

To test context dependence, we isolate target-span clips for the 46 confirmed masked MiMo cases. Full-context masking is established by a case-specific ASR route; all clips use main strict ASR, so the MiMo comparison is cross-route. Table~\ref{tab:isolation} shows that isolation re-exposes 18/46 errors. Section~4.5 provides a same-decoder Qwen control.

\begin{table}[H]
\centering
\caption{Context isolation on 46 confirmed cases. Full sentences use the original audit route; clips use main strict ASR.}
\label{tab:isolation}
\begin{tabular}{lrrrr}
\toprule
Condition & Exp. & Mask & No out. & Other \\
\midrule
Original full sentence & 0 & 46 & 0 & 0 \\
Rough 6s clip & 16 & 11 & 17 & 2 \\
Aligned clip & 18 & 12 & 13 & 3 \\
\bottomrule
\end{tabular}
\end{table}

A rough 6-second extraction exposes 16 cases. Whisper-small timestamp alignment exposes 18: 16 strong erroneous-reading and 2 partial unit-letter recoveries. Representative isolated transcripts include 二十一至十七 for \emph{21-17}, F杠三五 for \emph{F-35}, and 伊尔负七十六 for 伊尔-76; a full-context route had recovered each to a surface-correct form.

This is not proposed as a replacement metric. It is mechanism evidence: ASR can recover the local acoustic evidence in many confirmed masked cases. Rather, sentence context can push the transcript back toward the intended or conventional written form.

\subsection{Human and automatic evaluation}

The 200-case Raw/Structured comparison is a secondary oracle-style diagnostic, interpreted as an upper-bound sanity check rather than a deployable-system result. It asks whether making reading decisions explicit reduces risk-span errors perceived by listeners. Overall case-macro accuracy is 0.8889 for Raw and 0.9503 for Structured, a +0.0614 gain (95\% CI: [+0.0352, +0.0891]). The largest case-macro gain is on CDRD-entity cases: 0.7887 to 0.9146, +0.1259 (95\% CI: [+0.0728, +0.1807]). This supports the interpretation that many failures are reading-decision failures rather than generic acoustic defects, but it is not presented as a fair frontend comparison. On the 30-case IAA subset, pairwise exact agreement for correct counts ranges from 0.900 to 0.933 on Raw and from 0.833 to 0.900 on Structured; Pearson correlations range from 0.848 to 0.948.

ASR-roundtrip scores are useful but protocol-sensitive and should not be treated as ground truth. Across ASR protocols, Raw/Structured automatic scores range from 0.495/0.528 under default ASR to 0.728/0.805 under MiMo-V2-Omni strict ASR, while the Raw-to-Structured direction remains stable. In the MiMo strict-ASR 200-case matrix, Raw scores 0.6121 and Structured scores 0.7826. On the aligned 196-case subset, human labels exceed strict-ASR scores by +0.2745 for Raw (95\% CI: [+0.2184, +0.3320]) and +0.1667 for Structured (95\% CI: [+0.1146, +0.2220]), showing a calibration gap rather than a direct masking rate.

\subsection{Cross-system validation}

CosyVoice is evaluated with Raw input on the same 110 targeted pool, so it tests cross-TTS recurrence rather than prevalence. Human review follows the same audio-first protocol. CosyVoice yields 51 confirmed masked cases; MiMo strict ASR masks 37, Whisper-small masks 36, and both mask 22. Its 27 exposed TTS errors, versus 9 for MiMo, also show that error realizations are TTS-dependent.

Occurrence-aware review of the 97 files previously labeled confirmed masked gives the open-source ASR comparison in Table~\ref{tab:openasr}.

\begin{table}[H]
\centering
\caption{Occurrence-aware surface recovery on confirmed-masked files.}
\label{tab:openasr}
\small
\begin{tabular}{lrrr}
\toprule
ASR & MiMo & CosyVoice & Total \\
\midrule
Paraformer-zh & 0/46 & 2/51 & 2/97 \\
Qwen3-ASR-1.7B & 19/46 & 21/51 & 40/97 \\
\bottomrule
\end{tabular}
\end{table}

Paraformer preserves a wrong/noncanonical form in 88/97 cases. Its FunASR \texttt{use\_itn} toggle changes 0/220 full-file and 0/46 aligned-clip transcripts, so this is a no-effect check rather than an ITN ablation. Qwen preserves a wrong/noncanonical form in 56/97. Among its 19 full-context MiMo recoveries, aligned-span transcription re-exposes 12 and leaves 7 surface-correct. Thus an open-source ASR-LLM reproduces both masking and the isolation effect, while Paraformer shows strong decoder and protocol dependence.

\section{Discussion}

The central claim is deliberately narrow: ASR-roundtrip can miss reading errors perceived by listeners in CDRD and CDRD-adjacent spans. This does not imply that ASR always masks such errors, that 46/110 is a natural production rate, or that Structured is a deployable frontend. ASR-roundtrip remains useful for scalable screening and ablation, but for Chinese news reading-risk benchmarks it should not be treated as standalone ground truth. Human listening references, and span-level diagnostics when mechanism claims are made, remain necessary.

This has practical implications for speech and language processing evaluation. Short written forms carry language-model, normalization, and domain-convention priors for both TTS reading decisions and ASR transcript recovery. The masking route therefore differs across cases: score and entity examples mainly support contextual recovery, while unit and membership examples often involve written-form canonicalization. Evaluating only the transcript can still hide precisely the fluent wrong readings that matter to listeners. Future work should test more open-source ASR/TTS systems and broaden risk-span discovery beyond the current rules; this paper focuses on documenting the ASR-roundtrip blind spot.

The main limitations follow the same boundary: the 110-row audit is targeted rather than random; CosyVoice reuses that pool and is Raw-only; and isolated clips can produce no-output or other transcripts. These experiments establish mechanism and evaluator dependence, not production incidence or a deployable isolation metric.

\textbf{Data availability.} Released prompts, settings, transcripts, labels, metadata, audio, summaries, and scoring code are available on \href{https://github.com/Jayden-X-L/cn-newstts-asr-roundtrip-masking}{GitHub} and archived on \href{https://doi.org/10.5281/zenodo.21454402}{Zenodo}. Code is MIT; the 500 company-authorized scripts, 5K synthetic cases, and other released data are CC BY 4.0. The 108,124-item source export is excluded.

\clearpage
\section{Compliance with Ethical Standards}

This study uses news text, synthesized speech, ASR transcripts, and human listening judgments for TTS evaluation. Human annotators performed span-level listening audits of synthetic audio and authorized the use of their anonymized judgments for this study. No sensitive personal data, medical data, biometric identification task, speaker identification task, or clinical intervention is involved. The paper reports aggregate evaluation results and does not release annotator identities. The study follows the principles of the Declaration of Helsinki. It was reviewed under the authors' institutional legal and compliance process and was determined not to require formal ethics-board review.

\textbf{Funding and competing interests.} No external funding was received for this study. The authors are employees of NetEase Cloud Music. The authors declare no other relevant financial or non-financial competing interests.

\end{document}